# Learning from Sparse Data by Exploiting Monotonicity Constraints

**Eric E. Altendorf**
Oregon State University
Corvallis, OR 97331

**Angelo C. Restificar**
Oregon State University
Corvallis, OR 97331

**Thomas G. Dietterich**
Oregon State University
Corvallis, OR 97331

## Abstract

When training data is sparse, more domain knowledge must be incorporated into the learning algorithm in order to reduce the effective size of the hypothesis space. This paper builds on previous work in which knowledge about qualitative monotonicities was formally represented and incorporated into learning algorithms (e.g., Clark & Matwin's work with the CN2 rule learning algorithm). We show how to interpret knowledge of qualitative influences, and in particular of monotonicities, as constraints on probability distributions, and to incorporate this knowledge into Bayesian network learning algorithms. We show that this yields improved accuracy, particularly with very small training sets (e.g. less than 10 examples).

## 1 Introduction

Computer systems constructed with machine learning give the best known performance in many domains including speech recognition, optical character recognition, bioinformatics, biometrics, anomaly detection, and information extraction. However, it is not correct to view these systems as having been constructed purely from data. Rather, every practical machine learning system combines knowledge engineering with data. A critical and time-consuming part of building any successful machine learning application is the feature engineering, feature selection, and algorithm selection required to effectively incorporate domain knowledge.

One of the reasons this process is so time consuming is that machine learning tools do not provide very many ways of expressing domain knowledge. In particular, there are many forms of prior knowledge that an expert might have that cannot be accepted or exploited by existing machine learning systems. This paper discusses one particular form of prior knowledge—knowledge about qualitative monotonicities—and describes how this knowledge can be formalized and incorporated into learning algorithms for Bayesian networks.

Researchers in qualitative physics have developed several formal languages for representing qualitative influences [5, 16]. Others have shown that these qualitative influences could be usefully incorporated into learning algorithms, including the CN2 learning system, decision tree algorithms, and the back-propagation neural network algorithm [7, 2, 3, 9, 11, 8, 20, 15]. In this paper, we have chosen Bayesian network learning algorithms, because Bayesian networks already make it easy to express the causal structure and conditional independencies of a domain. We formalize qualitative monotonicities in terms of first-order stochastic dominance, building on the work of Wellman [27]. This in turn places inequality constraints on the Bayesian network parameters and leads naturally to an algorithm for finding the maximum likelihood values of the parameters subject to these constraints. Finally, we show experimentally that the additional constraint provided by qualitative monotonicities can improve the performance of Bayesian network classifiers, particularly on extremely small training sets.

## 2 Monotonicity Constraints

A monotonic influence, denoted $X \overset{Q+}{\succ} Y$ (or $X \overset{Q-}{\succ} Y$), informally means that higher values of $X$ stochastically result in higher (lower) values of $Y$. For example, we might expect a greater risk for diabetes in persons with a higher body mass index.

Our basic question, then, is: how does the statement $X \overset{Q+}{\succ} Y$ constrain a probability distribution $P(Y \mid X)$? Although there are various definitions of stochastic ordering [17, 24, 26], we employ *first order stochastic dominance* (FSD) monotonicity, which is based on the intuition that increasing values of $X$ shift the probability mass of $Y$ upwards. This leads to the following three definitions.

**Definition (First Order Stochastic Dominance)** *Given*

| | $P(Y \mid X)$ | | |
|---|---|---|---|
| $x$ \ $y$ | 0 | 1 | 2 |
| 0 | $\theta_0$ | $\theta_3$ | $\theta_{03}$ |
| 1 | $\theta_1$ | $\theta_4$ | $\theta_{14}$ |
| 2 | $\theta_2$ | $\theta_5$ | $\theta_{25}$ |

Constraints :

$\theta_0 \geq \theta_1$

$\theta_1 \geq \theta_2$

$\theta_0 + \theta_3 \geq \theta_1 + \theta_4$

$\theta_1 + \theta_4 \geq \theta_2 + \theta_5$

Figure 1: Example of a CPT for a three-valued variable $Y$ given a three-valued parent $X$, with constraint $X \overset{Q+}{\succ} Y$. The values for $\theta_{ab}$ are given by $1 - \theta_a - \theta_b$.

*two probability distributions $P_1$ and $P_2$, and their respective cumulative distribution functions $F_1$ and $F_2$,*

$$P_1 \ \succeq_{(1)} \ P_2 \quad \textit{iff} \quad \forall y \, F_1(y) \leq F_2(y). \tag{1}$$

**Definition (FSD Monotonicity)** *We say $Y$ is FSD isotonic (antitonic) in $X$ in a context $C$ if for all $x_1, x_2$ such that $x_1 \geq x_2$ (respectively, $x_1 \leq x_2$), we have*

$$P(Y \mid X = x_1, C) \ \succeq_{(1)} \ P(Y \mid X = x_2, C). \tag{2}$$

**Definition ( $\overset{Q+}{\succ}$ , $\overset{Q-}{\succ}$ Statements)** *Suppose $Y$ has multiple parents $X_1, X_2 \ldots X_q$. The statement $X_i \overset{Q+}{\succ}$ ( $\overset{Q-}{\succ}$ )$Y$ means for all contexts (configurations of other parents) $C \in \times_{j \neq i} X_j$, that $Y$ is FSD isotonic (antitonic) in $X_i$ in context $C$.*

The last definition expresses a *ceteris paribus* assumption, namely that the effect of $X$ on $Y$ holds for *each* configuration of other parents of $Y$.[1] A simple example of the induced monotonic CPT constraints is shown in Figure 1.

## 3 Inference and Learning

Our approach to learning with prior knowledge of monotonicities is to apply the monotonicity constraints to the parameter estimation problem to find the set of parameter values that gives the best fit to the training data while also satisfying the constraints. This is a form of constrained Maximum A Posteriori (MAP) estimation.

Let $G$ be the graph of a Bayesian network with nodes $X_1, \ldots, X_n$. Let $\theta_i$ denote the parameters of the CPT for $X_i$. Let $\theta_{ij}$ denote the row corresponding to parent configuration $j$, where $j \leq q_i$, the total number of parent configurations for $X_i$. Let $\theta_{ijk}$ denote the $k$th scalar in that vector, for $k \leq r_i$, the number of states of $X_i$. Finally, let $\theta$ denote the entire collection of parameters.

The learning task involves finding the most probable values for $\theta$ given our fully observed data $D$ and our prior $\xi$ over the parameters. In this case, $\xi$ is comprised of: (1) $\xi^G$, the conditional independence assumptions corresponding to the structure of $G$; (2) $\xi^Q$, the monotonicity constraints implied by our qualitative model $Q$; and (3) $\xi^P$, the prior over parameter values (e.g., a Dirichlet distribution for each conditional distribution). Under these priors, the likelihood for $\theta_i$ factors as

$$I_{(\theta_i \ satisf. \ \xi^Q)} \prod_{j=1}^{q_i} \prod_{k=1}^{r_i} \theta_{ijk}^{N_{ijk}}, \tag{3}$$

where $I_{(B)}$ is an indicator function that is 1 if $B$ is true and 0 otherwise (in this case, if $\theta_i$ satisfies constraints $\xi^Q$), and $N_{ijk}$ is the observed number of examples in which $X_i = k$ given parent configuration $j$, plus the appropriate parameter from a Dirichlet prior.

This is a constrained optimization problem which we solve with an exterior penalty method, replacing the indicator function in Equation 3 with penalty functions that take on large negative values when the constraints are violated. This approach is prone to problems with convergence, but it is flexible and scales linearly with the number of constraints, which can be very large in our problems.

### 3.1 Reparameterization and Notation

We first reparameterize the problem to eliminate the simplex constraints $\left(\sum_k \theta_{ijk} = 1\right)$. We define $\mu_{ijk}$ such that

$$\theta_{ijk} \equiv \frac{\exp(\mu_{ijk})}{\sum_{k'=1}^{r_i} \exp(\mu_{ijk'})}. \tag{4}$$

Although this adds a redundant free parameter, we had no problems with runaway optimization. For the following formulas, we introduce an abbreviated notation, defining $Z^i_{jk_c} \equiv \sum_{k=1}^{k_c} \exp(\mu_{ijk})$. Here and elsewhere, summation and product bounds over parent configurations and local states, when unspecified, are taken to be complete, that is, $1 \ldots q_i$ for $j$, and $1 \ldots r_i$ for $k$.

### 3.2 Likelihood Function

Our goal is to maximize, subject to the qualitative constraints, the likelihood $\prod_{jk} \left( \frac{\exp(\mu_{ijk})}{\sum_{k'} \exp(\mu_{ijk'})} \right)^{N_{ijk}}$. It is equivalent to maximize the natural logarithm,

$$J_L(\theta_i) \ = \ \sum_{jk} N_{ijk} \left( \mu_{ijk} - \ln \left( \sum_{k'} \exp(\mu_{ijk'}) \right) \right). \tag{5}$$

The gradient is defined by the partials

$$\frac{\partial}{\partial \mu_{ijk}} J_L(\theta_i) \quad = \quad N_{ijk} - \frac{\exp(\mu_{ijk})}{Z^i_j} \sum_{k'} N_{ijk'}. \tag{6}$$

[1] Kuipers [16] discusses $M+$ monotonicity, which has an unconditional positive effect *across all* configurations of the other parents (i.e., the monotonicity effect is global and cannot be overridden by other influences). Other notations for $\overset{Q+}{\succ}$ include $S+$ (Wellman [27]) and $\propto_{Q+}$ (Forbus [12]).

### 3.3 Inequality Constraint Margins

When observed data violates a monotonicity constraint, the maximum likelihood parameters are invariant to the parent configuration. It is debatable whether or not this is the intended or desired behavior. The proper solution would be a soft Bayesian prior on monotonicity which is updated with data, but for computational reasons we choose a simpler strategy. We enforce the strength of monotonicity by adding a margin to each inequality, replacing equation 1 by

$$P_1 \succeq_{(1)} P_2 \quad \text{iff} \quad \forall y\, F_1(y) + \varepsilon \leq F_2(y). \tag{7}$$

We must be careful not to make $\varepsilon$ too large, or it will strengthen the constraints to a point at which they have no solution (this is because inequalities are transitive, e.g.: $F_1(y)+\varepsilon \leq F_2(y), F_2(y)+\varepsilon \leq F_3(y), \ldots$). The maximum length of such a "chain" of inequalities is the Manhattan distance between the minimum-influence and maximum-influence corners of the CPT: $d_1^i = \left|\mathbf{pa}_i^{max} - \mathbf{pa}_i^{min}\right|_1 = \prod_{p\in\pi_i}(r_p - 1)$, where $\pi_i$ is the set of parents of $X_i$. For example, if $q_i = 2$, and each parent has 3 states, we get $d_1^i = 4$ inequalities, and our maximum allowable value for $\varepsilon$ is 0.25. Thus we define a global margin parameter $\varepsilon$ and let each node $X_i$ have its own $\varepsilon_i$ margin, where $\varepsilon_i = \varepsilon / d_1^i$. Theoretically, $\varepsilon$ could range up to 1.0, but we find that our current gradient search algorithms have difficulty finding the feasible region for $\varepsilon$ greater than 0.2.

### 3.4 Constraints and Penalty Functions

We now consider our exterior penalty functions. Individual inequality constraints resulting from monotonicity statements are indexed by four variables: the node $i$ to which the constraint applies, the two parent configurations $j_1, j_2$ being compared, and the state index $k_c$ for which the cumulative distribution function is evaluated ($c$ is not a variable, but only stands for "cumulative distribution function"). Without loss of generality, we consider monotonically increasing constraints ( $\overset{Q+}{\succ}$ ), such that the cdf corresponding to parent configuration $j_1$ is greater than or equal to the cdf for $j_2$, when $j_1 < j_2$. We denote a particular constraint as $C_{j_1,j_2}^{i,k_c}$. Thus, $\xi_i^Q \equiv \left\{C_{j_1,j_2}^{i,k_c} \mid j_1 < j_2 \leq q_i \wedge k_c < r_i\right\}$. We define the epsilon-margin modified constraint $C_{j_1,j_2}^{i,k_c}$ as

$$\begin{aligned} 0 &\geq P(X_i \leq k_c \mid \mathbf{pa}_i^{j_2}) - P(X_i \leq k_c \mid \mathbf{pa}_i^{j_1}) + \varepsilon \\ &= \sum_{k'=1}^{k_c} \theta_{ij_2k'} - \sum_{k'=1}^{k_c} \theta_{ij_1k'} + \varepsilon \\ &= \sum_{k'=1}^{k_c} \frac{\exp(\mu_{ij_2k'})}{Z_{j_2}^i} - \sum_{k'=1}^{k_c} \frac{\exp(\mu_{ij_1k'})}{Z_{j_1}^i} + \varepsilon \\ &= \frac{Z_{j_2k_c}^i}{Z_{j_2}^i} - \frac{Z_{j_1k_c}^i}{Z_{j_1}^i} + \varepsilon \qquad \equiv \delta. \end{aligned} \tag{8}$$

This term, denoted as $\delta$, will be positive when the constraint is violated. Thus, for each constraint $C_{j_1,j_2}^{i,k_c}$ we define the natural penalty function

$$P_{j_1,j_2}^{i,k_c} = I_{(\delta>0)}\ \delta^2. \tag{9}$$

The gradient is given by

$$\begin{aligned} \frac{\partial}{\partial\mu_{ijk}} P_{j_1,j_2}^{i,k_c} \quad = \quad & 2I_{(\delta\geq 0)}\ \delta \exp(\mu_{ijk}) \left(I_{(j=j_2)} - I_{(j=j_1)}\right) \\ & \left(\left(I_{(k\leq k_c)}\ Z_j^i - Z_{jk_c}^i\right) / \left(Z_j^i\right)^2\right). \end{aligned}$$

After applying the exterior penalty methods, the final function to optimize will be the log-likelihood minus the sum of the penalty functions times a penalty weight $w$:

$$J(\theta_i) = J_L(\theta_i) - w \sum_{C_{j_1,j_2}^{i,k_c} \in \xi_i^Q} P_{j_1,j_2}^{i,k_c}. \tag{10}$$

## 4 Experiments and Evaluation

To test the effectiveness of qualitative monotonicities, we conducted a series of experiments comparing Bayesian network classifiers learned with and without qualitative monotonicities.

### 4.1 Data Sets

We have chosen five data sets from the UCI ML repository: auto-mpg[22], haberman[14], pima-indian-diabetes[23], breast-cancer-wisconsin[4], and car[6, 30]. For each of these data sets we constructed the network (KB structure) using domain knowledge, and inserted monotonicity annotations ( $\overset{Q+}{\succ}$ or $\overset{Q-}{\succ}$ ) on each of the network links according to our domain knowledge. This domain knowledge was based only on common knowledge (e.g., car purchasing) and information from previous publications concerning these data sets. In particular, we did not examine the data itself. We invested significant research in this task, and consider our constructed models the beginning of a benchmark corpus for monotonic learning algorithms.

We had originally chosen ten datasets, but of these, only five had a tractable Bayes net structure. The others had nodes with 8–11 incoming arcs, making the optimization task very difficult, and yielding low performance on all Bayes net classifiers (since we are relying on our domain knowledge, we chose to drop these datasets rather than modify the networks in a way that disagreed with our causal understanding of the domain).

We hypothesized that monotonicity constraints would prove more helpful at finer discretizations. To test this, for each data set, attributes with numeric values were discretized using Weka's (the Waikato Environment for

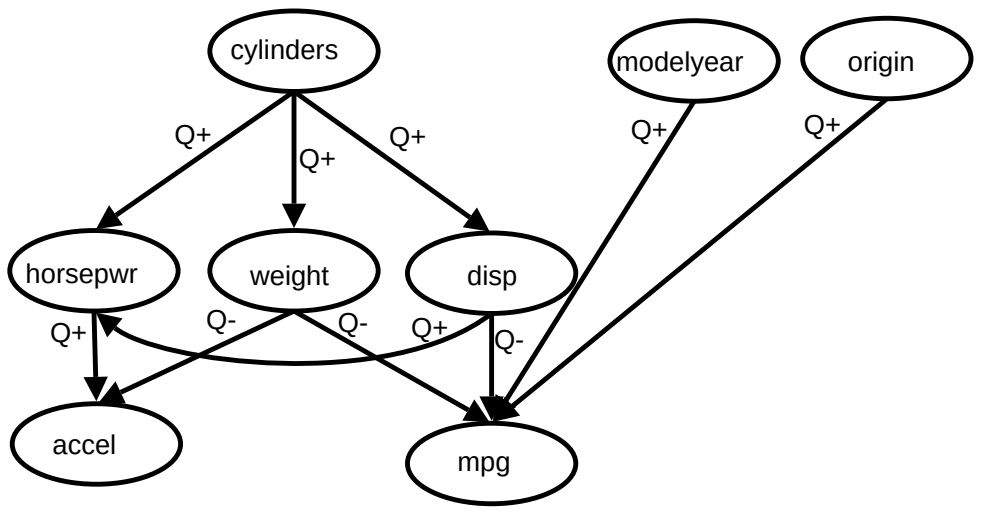

Figure 2: auto-mpg

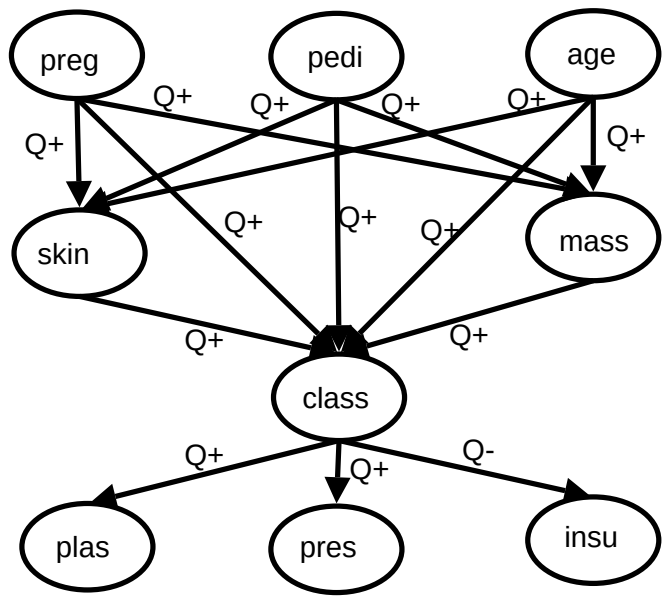

Figure 3: pima-indian-diabetes

Knowledge Analysis [29]) equal-frequency discretization tool to generate data sets with numbers of bins 2, 3, and 5, yielding a total of 15 data sets for our experiments. All class variables have two classes. Moreover, all incomplete rows in the data sets have been removed.

Figure 2 shows the KB structure and monotonicity constraints for data set auto-mpg. In this data set, the classification problem is to predict whether a car has low ($\leq 28$) or high ($> 28$) mileage per gallon (mpg). auto-mpg has 392 instances of which 106 are labeled positive examples. Domain knowledge suggests that an increase in the number of cylinders (cylinders) usually leads to an increase in horsepower (horsepwr), displacement (disp), and vehicle weight (weight). An increase in weight leads to a decrease in mpg. The heavier the vehicle, the slower it accelerates (accel). The larger the displacement, the greater the horsepower, but the lower the mileage per gallon. Finally, newer models (modelyear) tend to be more fuel-efficient, as do vehicles imported from (origin) Japan (encoded as 1) as opposed to Europe or those produced in the United States (encoded as 0). These monotonicity relations are encoded as constraints in the network as shown in Figure 2.

Figure 3 shows the KB structure and monotonicity constraints for pima-indian-diabetes. The problem for this data set is to classify examples that tested positive for diabetes. This data set has 768 instances of which 268 are labeled positive. Domain knowledge suggests that an increase in each of the triceps' skin fold thickness (skin) is expected with an increase in the number of experienced pregnancies (preg), an increase in age (age), and perceived risk due to pedigree (pedi). The same monotonic relations are also suggested in body mass index (mass). An increase in preg, pedi, age, skin, or mass increases the risk of diabetes (class). Most diabetics have high levels of plasma glucose concentration (plas) and most suffer from high blood pressure (pres) while having low levels of insulin (insu).

The KB structure and monotonicity constraints for breast-cancer-wisconsin are shown in Figure 4. The classification problem is to predict whether a given example is malignant (malignant) or benign. The data set breast-cancer-wisconsin has 683 examples, of which 239 are positive. The attributes in the database have been assigned values that range from 1 (normal state) to 10 (most abnormal state). The attributes are: clump thickness (clumpthick), uniformity of cell size (cellsize), uniformity of cell shape (cellshape), single epithelial cell size (epitsize), bare nuclei (barenuc), normal nucleoli (normnuc), mitoses, marginal adhesion (adhesion), and bland chromatin (blandchr). These attributes have been visually assessed using fine needle aspirates taken from patients' breasts. Malignant samples have observed abnormal states, i.e., the more the malignant a sample the higher the state of abnormality. Hence, all network links from malignant to other attributes have $\overset{Q+}{\succ}$ monotonicity constraints.

Figure 5 shows the KB structure and monotonicity constraints for haberman. The problem in this data set is to predict the survival status of a patient who has undergone breast cancer surgery. haberman has 306 instances of which 225 are positive examples. The data set has three attributes: the age of patient at time of operation (age), the patient's year of operation (year), and the number of positive axillary lymph nodes detected (nodes). We expect the survivability of the patient to decrease as the patient gets older, to decrease as the number of positive nodes detected increases, and increase with the operation year, i.e., more recent implying better survival.

Figure 6 shows the relevant information for the car data set. The prediction problem in this data set is to determine whether a given instance is acceptable (class) given the following attributes: (price), cost of maintenance (maint), capacity in number of persons (person), size of luggage space (luggage), estimated safety rating (safety), and number of doors (doors). car has a total of 1728 examples of which 30% are positive. This data set has non-numeric attributes which we re-encoded as numbers, in their respective ordinal order. Common knowledge suggests that as price and maintenance costs increase, acceptability should decrease. Increases in the safety rating and passenger capacity would generally increase acceptability. We assume the number of doors does not play a significant role in a car's acceptability. Also, we may expect that an increase in a car's safety, passenger capacity, or luggage space could lead to an increase in price. These monotonicity relations are encoded in the network shown in Figure 6.

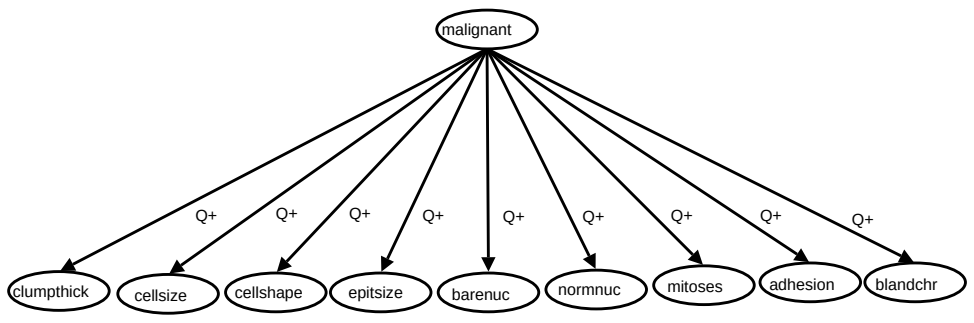

Figure 4: breast-cancer-wisconsin

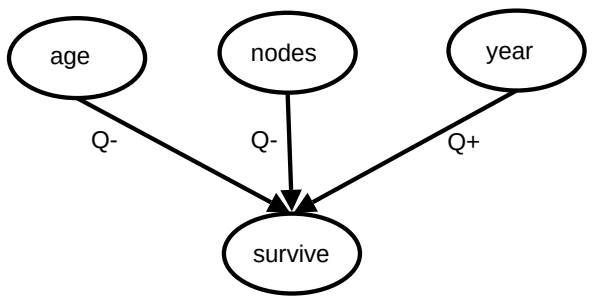

Figure 5: haberman

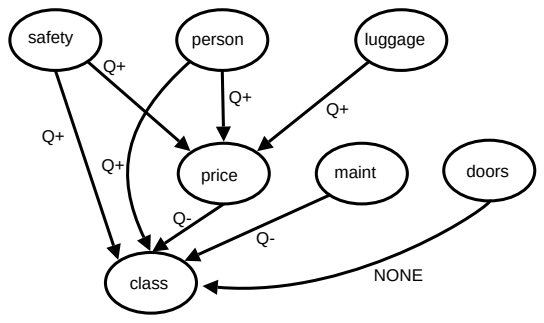

Figure 6: car

### 4.2 Experimental Setup

We selected an implementation of the L-BFGS[2] algorithm from the conditional random fields package Mallet [18] to optimize our penalized objective function. To ensure convergence in the feasible region, the L-BFGS maximization was wrapped in the outer-level algorithm given in Figure 7.

1. Initialize the $\mu_{ijk}$ parameters at the unconstrained MLE point (found simply by counting the observations)
2. If this point satisfi es the constraints, return it
3. Otherwise, initialize a weight $w$ for the penalty functions
4. Take steps in the steepest direction of the penalized likelihood until convergence
5. If we converged outside the feasible region, increase the penalty weight and repeat the previous step.

Figure 7: Constrained optimization algorithm

In addition, L-BFGS would sometimes fail to converge, or simply fail to continue from certain points in the parameter space. We suspect that these problems were due to the relatively sharp edges at the constraint boundary. To work around this, upon failure we slightly increased the penalty weight $w$ (just to change the shape of the function) and re-ran the maximizer routine (perhaps many times). This proved sufficient for many of our experiments, though we still had to be careful not to set our $\varepsilon$ margin too high, since it would make the feasible region so small that it might never be found. We also experimented with variations—for example, using cubed violations for penalty functions rather than squared, or using a very small exponent (such as 1.01). None of the variations resulted in significantly more reliable estimation routines.

For running experiments, we integrated the learning algorithm with Weka, which allowed us to easily script learning runs and to run comparisons against other learning algorithms. The algorithms we analyzed were:

**Zero-Regression (ZR)** Always picks the mode of the observed distribution of the class variable, without regard to the features.

**Naïve Bayes (NB)** Also known as the simple Bayesian classifier (SBC). Treats the class variable as the parent in a Bayesian network, with all features as children.

**Knowledge-based Bayes (KB)** Fit the parameters of a Bayesian network whose structure incorporates domain knowledge—specifically, the Bayesian network structures shown in Figures 2-6. Parameters are fit by maximum likelihood with a Laplace correction.

**Constrained Knowledge-based Bayes (CKB)** Same as KB, except that the parameters are fit to maximize the posterior probability subject to the inequality constraints induced by the qualitative monotonicity statements. CKB was run with three different margins, $\varepsilon \in \{0.0, 0.1, 0.2\}$. These runs are designated CKB0, CKB0.1, and CKB0.2.

[2]Limited-memory BFGS, a variation of the the *Broyden-Fletcher-Goldfarb-Shanno* algorithm (see, e.g., [21, pg.324]).

### 4.3 Results

To compare the algorithms on each data set, we first randomly split the data set into a test set (1/3 of the data) and a training set pool (2/3 of the data), stratified by class. Then we performed 50 replications for each training set size $m$, for various $m$ from 1 to 50. In each replication, we randomly drew $m$ elements without replacement from the training set pool, and trained our algorithms on the set. The resulting fitted networks were then evaluated on the test set. The results are shown in Figure 8. We also present in Figure 9 the results of running McNemar's test on the NB, KB, CKB0 and CKB0.1 classifiers trained on the pima data set.

Let us begin by considering the performance of the different algorithms on very small samples. Our hypothesis was that ZR would perform the worst, that Naïve Bayes would be the second worst, that the Knowledge-based networks would come next, and that the networks that combined knowledge-based structure with monotonicity constraints (CKB) would give the best results. The plots show that actually, ZR performs surprisingly well: comparable to or better than NB at small sample sizes on all data sets. Furthermore, on haberman, NB and ZR dominate, even on small samples. Otherwise, we do see the expected ranking, though at small sample sizes, ZR and NB frequently tie, as do KB and CKB.

Since our largest training set was of size 50, which we still consider relatively small for models of the complexity used here, we actually expected to see this ranking extend

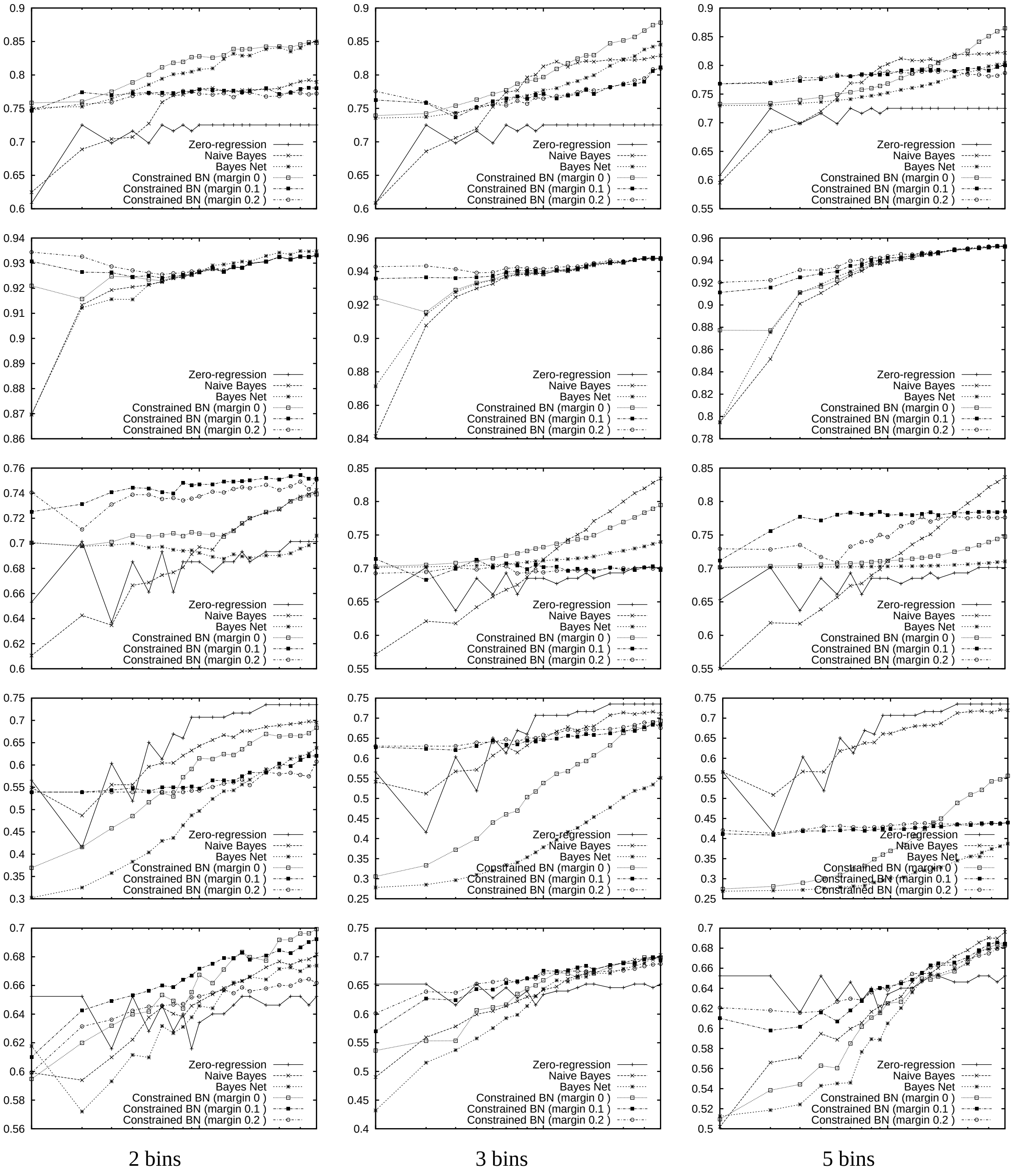

Figure 8: Learning curves for auto, bcw, car, haberman, and pima domains at 3 discretizations, plotting average accuracy (across 50 runs) against training set size (log scale, 1 through 50). Zero-regression is omitted on the bcw data set as it had performance far below the other algorithms.

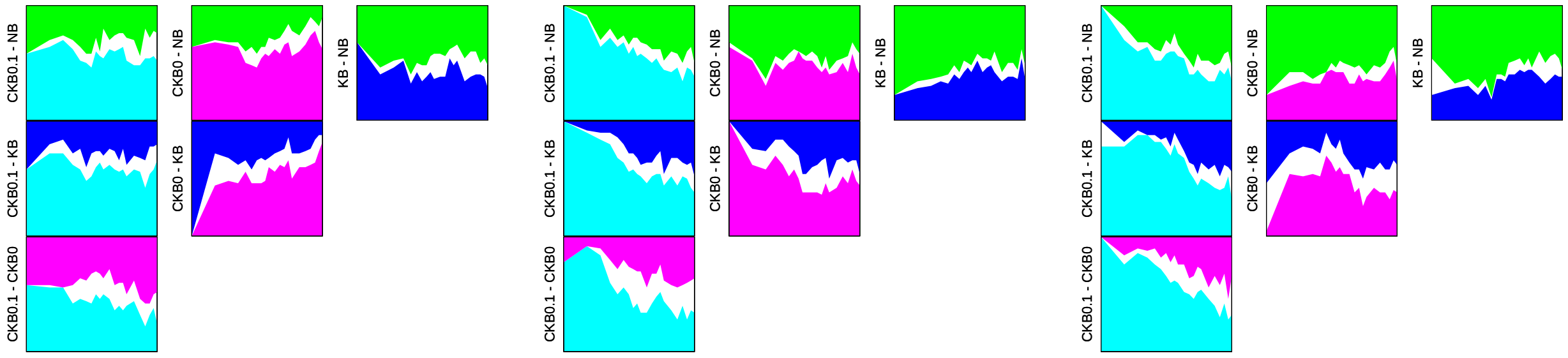

2 bins 3 bins 5 bins

Figure 9: McNemar's test for pima at 2, 3, and 5 bins, comparing pairs of algorithm from CKB0.1, NB, CKB0, and KB (in order from lightest to darkest shade). The 6 pairwise comparisons are run at training set sizes 1 through 50 ($x$ axis, log scale). The lower and upper regions represent the number of statistically significant wins of each algorithm, with the remaining center region indicating ties or statistically insignificant wins.

through more of the tested sample sizes. One somewhat surprising result was how well NB performed on car at higher sample sizes with discretizations finer than 2 bins. We were also particularly surprised by the results on the haberman data set, where NB and ZR did very well all the way through $m = 50$. A simple data analysis on the haberman data (2-bin) using correlation and mutual information revealed that the data set exhibits independence between the class variable and any one of the three parent attributes. Moreover, the conditional probability tables reveal that the parameters do not exhibit monotonicity; for example, the chance of surviving given that the patient is young is high but surprisingly the data also says that the chances of surviving given that the patient is old is also high. In this case, our assumptions about the structure and monotonicities were clearly incorrect.

A second hypothesis was that the monotonicity constraints might be incorrect and lead to poor performance at large sample sizes, particularly with $\varepsilon = 0.2$. The plots do show flatter learning curves for CKB with $\varepsilon > 0$, compared to CKB with no margin, and this difference is also clearly shown in the McNemar's test comparisons between the two. However, without margins, CKB is comparable to or better than KB at nearly all tested sample sizes.

A third hypothesis, as mentioned, was that monotonicity constraints would help more at finer discretizations. The plots show some support for this; on auto, there is little difference between CKB0 and KB at the 2-bin discretization, and at higher discretizations and with more training data, CKB0 dominates KB. CKB0.1 on auto shows very good performance at the 5-bin discretization level. On bcw, all algorithms perform about the same at high sample sizes, though CKB0.1 does well at low sample sizes, and this effect is amplified at finer discretizations. Finally, looking only at lower sample sizes, we observe this effect on the pima data set. The results from the remaining data sets do little to support or disprove this hypothesis.

## 5 Related Work

Knowledge-based model construction employs domain knowledge to construct a model customized for a specific truth, probability, or decision query [28, pg.26]. The models are usually graphical (e.g., qualitative probabilistic networks [28] or dynamic belief networks [13, 19]), but generally do not incorporate learning [28, pg.31].

Clark and Matwin [7] share our motivations for knowledge-constrained learning, but perform learning only of parameters of a qualitative process model (temporal simulation).

Several researchers have incorporated prior knowledge into artificial neural networks. Knowledge-based ANN's [25] have structures and parameters initialized from knowledge bases of Horn clauses. ANN's also lend themselves to connection weight constraints that enforce monotonicity; such models are developed and applied in [2, 9, 15].

In [8], Daniels, Feelders, and Velikova improve on prior work in monotonic regression techniques by employing both constrained neural networks and decision trees. Monotonicity in trees in further discussed in [20, 11, 3]. Monotonicity in trees is important since CART and C4.5 tend to build nonmonotonic trees even given monotonic data [20]. Interestingly, some of the decision tree work (e.g., [11]) reports poorer classification performance under monotonicity constraints, the advantage being found in simpler, more understandable decision trees. Much of this work was also motivated by the need to make justifiable decisions (e.g., in school admissions or job or loan applications, see for instance [3]).

The literature on machine learning with monotonicities appears to be restricted to classifiers and regression models, despite the fact that the statistics and financial literature is rich with discussions of stochastic ordering. A seminal work is [17]; a comprehensive recent treatment can be found in [24]. The specific problem of estimating a pair of distributions $P_1(X)$ and $P_2(X)$ under the con-straint $P_1 \succeq_{(1)} P_2$ is discussed in [10], though we found no work on estimating partially ordered sets of distributions. Agresti and Chuang [1] discuss a stricter form of monotonicity defined on joint distributions by the local odds-ratios (with both a hard sign constraint and a Gaussian prior), and present techniques for both constrained MLE and MAP inference under such priors.

## 6 Summary and Future Work

We have presented a method for using qualitative domain knowledge in the estimation of parameters for a probabilistic model. The qualitative statements (monotonicity) are natural and easily specified by domain experts, and they have a probabilistic semantics consistent with a domain expert's intuition. We have shown that these semantics are tractable and can effectively constrain probability distributions, and that the constrained models in many cases generalize from the training set to the test set better than do unconstrained models.

In other work, we have defined a language for expressing qualitative constraints, including monotonicity, synergy, strength of influence, and saturation, all defined in terms of constraints on conditional probability distributions. We plan on testing the efficacy of these other qualitative constraints in different domains. We also will research more efficient and reliable optimization methods. Finally, we wish to develop monotonic constraints in simplified interaction models (such as noisy-or) and test on the five rejected datasets whose KB structure had proved intractable for Bayes net classifiers in this work.